\documentclass[10pt,twocolumn,letterpaper]{article}
\usepackage[pagenumbers]{iccv} 

\usepackage{multirow}
\definecolor{iccvblue}{rgb}{0.21,0.49,0.74}
\usepackage[pagebackref,breaklinks,colorlinks,allcolors=iccvblue]{hyperref}

\def\paperID{14479} 
\def\confName{ICCV}
\def\confYear{2025}

\title{PS3: A Multimodal Transformer Integrating Pathology Reports with Histology Images and Biological Pathways for Cancer Survival Prediction \thanks{Accepted at ICCV 2025}}

\author{Manahil Raza \quad Ayesha Azam \quad Talha Qaiser \quad Nasir Rajpoot \\
University of Warwick, UK\\
{\tt\small \{manahil.raza, ayesha.azam, talha.qaiser, n.m.rajpoot\}@warwick.ac.uk}}

\begin{document}
\maketitle
\begin{abstract}
Current multimodal fusion approaches in computational oncology primarily focus on integrating multi-gigapixel histology whole slide images (WSIs) with genomic or transcriptomic data, demonstrating improved survival prediction. We hypothesize that incorporating pathology reports can further enhance prognostic performance. Pathology reports, as essential components of clinical workflows, offer readily available complementary information by summarizing histopathological findings and integrating expert interpretations and clinical context. However, fusing these modalities poses challenges due to their heterogeneous nature. WSIs are high-dimensional, each containing several billion pixels, whereas pathology reports consist of concise text summaries of varying lengths, leading to potential modality imbalance. To address this, we propose a prototype-based approach to generate balanced representations, which are then integrated using a Transformer-based fusion model for survival prediction that we term PS3 (Predicting Survival from Three Modalities). Specifically, we present: (1) Diagnostic prototypes from pathology reports, leveraging self-attention to extract diagnostically relevant sections and standardize text representation; (2) Histological prototypes to compactly represent key morphological patterns in WSIs; and (3) Biological pathway prototypes to encode transcriptomic expressions, accurately capturing cellular functions. PS3, the three-modal transformer model, processes the resulting prototype-based multimodal tokens and models intra-modal and cross-modal interactions across pathology reports, WSIs and transcriptomic data. The proposed model outperforms state-of-the-art methods when evaluated against clinical, unimodal and multimodal baselines on six datasets from The Cancer Genome Atlas (TCGA). The code is available at: \url{https://github.com/manahilr/PS3}.
\end{abstract}    

\section{Introduction}
Patient prognosis in oncology refers to the predicted progression and outcome of cancer, often determined by clinical, pathological and molecular factors \cite{yasui2005molecular,pugliese2016clinical}. It is essential to guide treatment strategies, inform risk assessment and facilitate patient-centered decision-making to optimize survival and quality of life \cite{gill2012central,mackillop2003importance}. Recently, integrating multiple data modalities have been shown to improve survival prediction in cancer research, with most studies focusing on combining genomic data such as transcriptomic data with histology whole slide images (WSIs) \cite{stahlschmidt2022multimodal,steyaert2023multimodal,deepa2022systematic}.
These approaches leverage the strengths of each modality—genomics captures molecular subtypes, while WSIs provide spatial and morphological insights into tumor characteristics \cite{zhang2024prototypical}. However, pathology reports remain an under-utilized resource despite providing essential information to clinicians for estimating the prognosis of cancer patients. Generated by expert pathologists,  they are mandatory for definitive cancer diagnosis \cite{pena2009does} and contain critical clinical and prognostic information including biomarker status, tumor grading, staging and histological subtypes \cite{lee2018automated,napolitano2016machine,kefeli2024generalizable}. They often complement histology and genomic data by incorporating expert-driven diagnostic insights \cite{steimetz2024forgotten,abedian2021automated}.  Moreover, as routinely produced components of clinical workflows, pathological reports serve as a readily available source of medical knowledge \cite{kefeli2024tcga}. Despite their clinical significance, they remain largely untapped in computational survival prediction models.

Early multimodal approaches often relied on late fusion techniques  \cite{chen2022pan,li2022hfbsurv,wang2021gpdbn}, which integrated unimodal representations only at the final decision stage \cite{gadzicki2020early}. However, these methods struggled to model cross-modal interactions, which may be crucial for accurate prognosis. In contrast, more recent studies have investigated early fusion approaches for integrating WSIs and genomic data. Although early fusion captures fine-grained cross-modal interactions more effectively \cite{songmultimodal,jaume2024modeling,chen2021multimodal,xu2023multimodal}, it remains computationally demanding due to the high dimensionality and heterogeneity of the data. 

Another key challenge in early fusion of WSIs, genomics and pathology reports is modality imbalance, arising from differences in structure and scale of the data modalities.  WSIs are gigapixel-scale images typically divided into thousands of high-dimensional patches containing spatial and morphological information \cite{raza2024dual}. To process these vast amounts of image data, Multiple Instance Learning (MIL) \cite{gadermayr2024multiple,wang2024advances} is commonly employed, allowing models to extract and aggregate relevant features from individual patches. 
In contrast, pathology reports provide concise textual summaries with significantly fewer raw data points.  Further, genomic data, such as RNA-Seq expression levels are typically represented as isolated scalar values, lacking contextual information about their biological functions. 
This imbalance can lead to modality dominance, where data-rich modalities (\eg WSIs) can disproportionately influence the model compared to modalities with fewer tokens (\eg pathology reports). Addressing this imbalance is crucial for developing an effective multimodal fusion framework that ensures fair contributions from all modalities and improves predictive performance.

To address these challenges, we propose a prototype-based multimodal fusion framework that standardizes and balances representations across three modalities. The prototypes represent large, variable-length inputs as compact embeddings, enabling more effective multimodal fusion. We exploit the inherent morphological redundancy in WSIs by identifying recurring histological patterns within tissue patches \cite{vu2023handcrafted}. Inspired by previous works \cite{song2024morphological,songmultimodal,kim2022differentiable}, we use a Gaussian Mixture Model (GMM) to represent slides, with each mixture component corresponding to a distinct \textit{histological prototype}. This clustering of similar morphological patterns enables compact WSI representations using key histological prototypes. Pathology reports, which contain significantly less raw data than WSIs, often exhibit unstructured or semi-structured formats, inconsistent formatting, varying lengths and levels of detail \cite{napolitano2016machine,truhn2024extracting,santos2022automatic}. Rather than truncating their content, we construct \textit{diagnostic prototypes}—standardized representations designed to capture the most diagnostically relevant sections of each report using self-attention-based mechanisms \cite{vaswani2017attention}. To ensure a biologically meaningful representation of transcriptomic data, we draw inspiration from prior methods \cite{jaume2024modeling,zhang2024prototypical,songmultimodal} and transform the values into a set of 50 Cancer Hallmark \textit{pathway prototypes} \cite{liberzon2015molecular}. This representation aligns gene expression patterns with well-established cellular processes, providing a structured and interpretable genomic feature set.  

We propose a multimodal transformer-based framework \textbf{PS3}, designed to \textbf{P}redict \textbf{S}urvival by integrating information from three (\textbf{3}) different data modalities: WSIs, transcriptomics and pathology reports. By transforming raw data into compact prototype-based representations, we reduce the disparity in token counts across modalities, achieving a more balanced token distribution. This allows our transformer-based method to effectively model self-attention and cross-attention mechanisms, thereby facilitating the fusion of complementary information from all three modalities. As a result, our approach enhances predictive performance, leading to robust survival prediction. We evaluate the proposed method on six cancer cohorts from The Cancer Genome Atlas (TCGA) \cite{TCGA} and demonstrate that it outperforms both unimodal and multimodal baselines.

To summarize, our main contributions are: $(1)$ a novel framework for representing pathology reports, WSIs and transcriptomic profiles using diagnostic, histological and pathway prototypes, respectively; $(2)$ development of a multimodal transformer, PS3, to predict survival using three modalities; and $(3)$ comprehensive experiments and ablation studies on six cancer cohorts, showcasing the effectiveness of PS3 for cancer survival prediction.


\section{Related Work}
\subsection{Survival Prediction with Whole Slide Images}
Given the large number of patches in WSIs, many methods adopt MIL-based methods for efficient processing and analysis  \cite{afonso2024multiple}. MIL typically involves three key steps: (1) dividing the WSI into hundreds or thousands of smaller patches (or instances), (2) extracting patch embeddings using a feature encoder and (3) aggregating the patch embeddings to obtain a slide-level representation \cite{wang2024advances}. For survival prediction, various MIL-based approaches have been explored, including graph-based \cite{li2018graph,chen2021whole,liu2023graphlsurv} and attention-based MIL models \cite{ilse2018attention,yao2020whole} both of which aim to capture global WSI representations for survival analysis.  More recently, transformer-based architectures leveraging self-attention mechanisms \cite{vaswani2017attention,dosovitskiy2020image,jiang2023mhattnsurv} have been introduced, including hierarchical transformer architectures designed to capture multi-scale WSI representations \cite{shao2023hvtsurv,yan2023sparse}.


\subsection{Multimodal Survival Prediction}
While unimodal methods have demonstrated strong prognostic capabilities, multimodal approaches have led to further improvements in survival prediction. Most existing multimodal fusion methods integrate genomic or transcriptomic data with multi-gigapixel histology WSIs \cite{chen2022pan}. Early studies primarily relied on late fusion using techniques such as concatenation \cite{mobadersany2018predicting}, Kronecker product \cite{chen2020pathomic,wang2021gpdbn} or factorized bilinear pooling \cite{li2022hfbsurv,qiu2023deep}. However, since late fusion merges unimodal representations only at the final stage \cite{gadzicki2020early}, it fails to capture cross-modal interactions, leading to suboptimal integration. Similarly, text-based fusion using concatenation \cite{huang2020fusion} and weighted sum \cite{naidoo2024survivmil} to combine patient health records and medical images suffer from similar limitations. 

In contrast, early fusion approaches have gained significant attention for their effectiveness in modeling cross-modal interactions. Many models utilize transformers \cite{xu2023multimodalsurvey}, incorporating co-attention \cite{chen2021multimodal,zhou2023cross,liu2023mgct} and cross-attention mechanisms \cite{ding2023pathology,jaume2024modeling,songmultimodal,lv2022transsurv} to enhance feature integration. Some extend transformer architectures with hierarchical transformers
 \cite{li2022hierarchical} while others employ prototyping techniques to reduce data dimensionality before fusion, improving efficiency and minimizing the number of tokens processed \cite{song2024morphological}. Additionally, other approaches such as optimal transport-based methods \cite{xu2023multimodal,songmultimodal}, information bottle-necking techniques \cite{zhang2024prototypical} and graph-based methods \cite{zheng2024graph} have been explored as alternatives to transformers.


\section{Methodology}
We introduce PS3, a prototyping-based multimodal framework for survival prediction that integrates histology images, transcriptomic data and pathology reports. We explain the construction of the diagnostic prototypes, histological prototypes and pathway prototypes in Sections \ref{section: text prototypes}, \ref{section: histology prototypes} and \ref{section: pathway prototypes}, respectively. Section \ref{section: fusion} describes the Transformer-based multimodal approach, while Section \ref{section: post processing} outlines the post-fusion processing steps taken for survival prediction.

\begin{figure*}[ht]
\includegraphics[width=\linewidth]{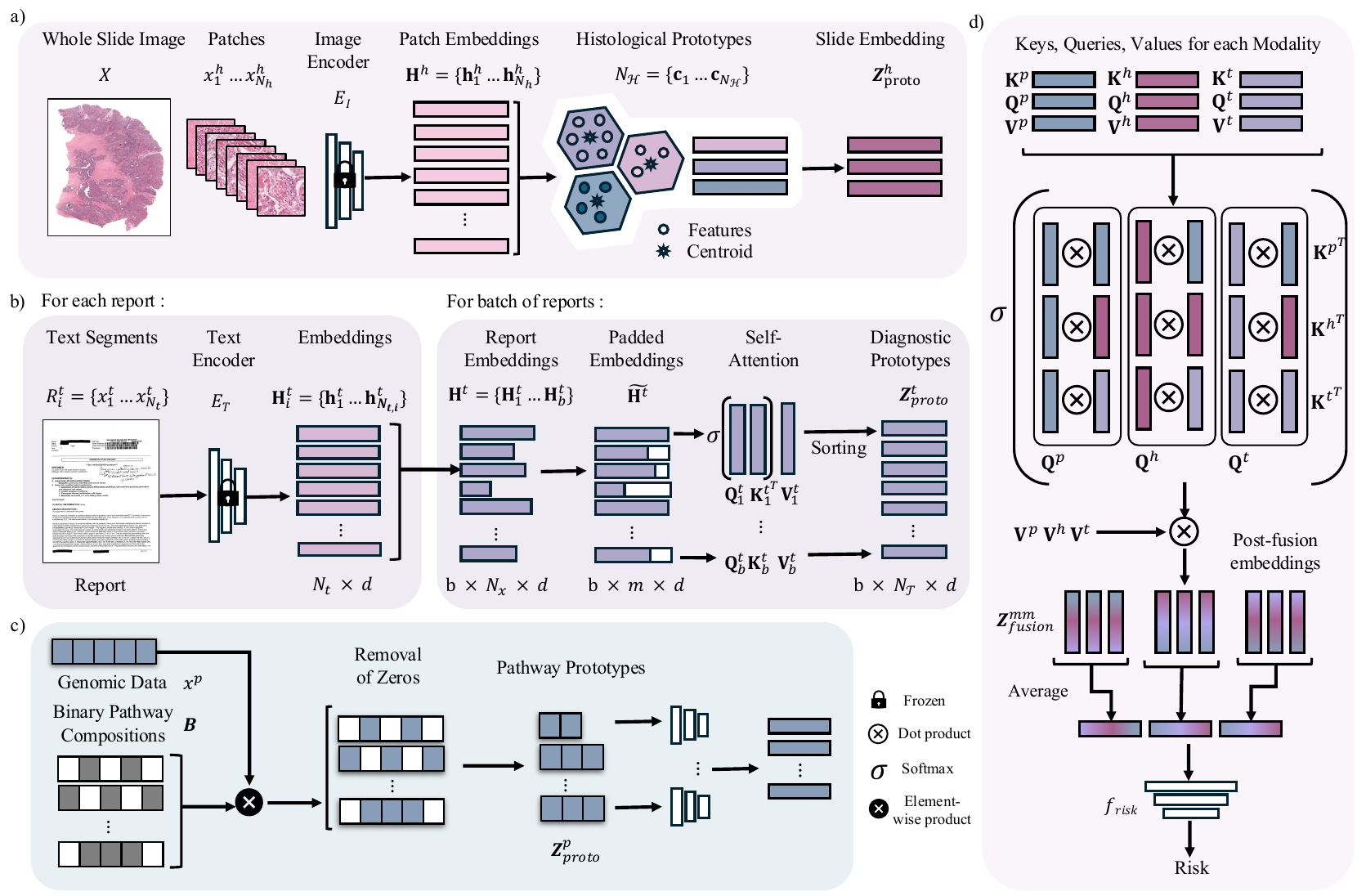}
\caption{(a) Whole Slide Image $X_i$ patches are processed by a frozen image encoder $E_I$ to extract visual features $\textbf{H}^h$, which are clustered into $N_\mathcal{H}$ histological prototypes using a Gaussian Mixture Model (GMM) to obtain a slide-level representation. 
(b) Feature embeddings $\mathbf{H}_i^t$ are extracted from the text segments of a single pathology report $R_i$, using a frozen text encoder $E_T$. For a batch of pathology reports $\mathbf{H}^t$ we construct Diagnostic prototypes by applying Transformer-based self-attention mechanisms and sorting the text segments within each report based on their importance scores. 
(c) Gene-expression vector $\mathbf{x}^p$ are multiplied with the binary pathway vectors ($\mathbf{B}$) to create pathway prototypes. These are processed through $f_{\alpha}$ MLPs to generate feature embeddings. (d) Prototypes from each modality are projected into key, value and query representations, followed by self- and cross-modal attention to integrate information. The resulting outputs are used for survival risk prediction.}
\label{main_fig}
\end{figure*}

\subsection{Diagnostic Prototypes from Pathology Reports}
\label{section: text prototypes}
\textbf{Feature Extraction: }Each pathology report ($t$), denoted as $R_i$, is first divided into smaller text segments or sections: $R_i = \{x_1^t,x_2^t \ldots x_{N_{t,i}}^t\}$, where ${N_{t,i}}$ is the number of segments in the $i^{th}$ report.  These text segments are tokenized and subsequently passed through a pre-trained and frozen text encoder $E_T$ which extracts feature embeddings for each text segment: $\mathbf{h}_i^t = E_T(x_i^t) \in \mathbb{R}^{d_t}$. The set of embeddings for the $i^{th}$ pathology report are then defined as: $\mathbf{H}_i^t = \{\mathbf{h}_1^t, \mathbf{h}_2^t \ldots \mathbf{h}_{N_{t,i}}^t\}$ .

\textbf{Self-Attention Mechanism: }For a batch of $b$ pathology reports, the input can be modeled as
$\mathbf{H}^t = \{\mathbf{H}_1^t, \mathbf{H}_2^t \ldots \mathbf{H}_b^t\} \in \mathbb{R}^{b \times N_{t,i}\times d}$ where ${N_{t,i}}$ is the number of text segments in the $i^{th}$ report. Since reports vary in length, they are padded to ensure uniformity within the batch. The resulting padded batch is represented as:
$\tilde{\mathbf{H}}^t = \{\tilde{\mathbf{H}}_1^t, \tilde{\mathbf{H}}_2^t \ldots \tilde{\mathbf{H}}_b^t\}\in \mathbb{R}^{b \times m \times d_t}$
where $m$ is the length of the longest report in the training set. A corresponding binary mask $ \mathcal{M} \in \{0,1\}^{b \times m}$ is created to differentiate between the real tokens and padded positions, ensuring they do not contribute to the attention scores.

We aim to standardize report representations while preserving the most clinically and pathologically relevant information. To achieve this, we employ a Transformer-based self-attention mechanism \cite{vaswani2017attention} to construct the Diagnostic Prototypes. In particular, we project the padded report embeddings into query, key and value vectors using learnable linear transformations:
\begin{equation}
    \mathbf{Q}^t = \mathbf{W}_Q \cdot \tilde{\mathbf{H}}^t, \quad \mathbf{K}^t = \mathbf{W}_K \cdot \tilde{\mathbf{H}}^t, \quad \mathbf{V}^t = \mathbf{W}_V\cdot \tilde{\mathbf{H}}^t
\end{equation}
where $\mathbf{W}_Q, \mathbf{W}_K, \mathbf{W}_V \in \mathbb{R}^{d_t \times d_t}$ are learnable matrices. The scaled dot-product attention $A^t$ is then computed as:
\begin{equation}
    \mathbf{A}^t = \sigma \left( \frac{\mathbf{Q}^t {\mathbf{K}^t}^\top}{\sqrt{d_t}} \right)
\end{equation}
The padded positions are masked during the self-attention process to prevent them from influencing the results. The attention weights are further used to compute a weighted sum of the values $\mathbf{Z}^t = \mathbf{A}^t 
\cdot \mathbf{V}^t \in \mathbb{R}^{b \times m \times d_t}$ representing the post-attention embeddings of the pathology reports. 

\textbf{Prototype Selection: }To extract the most diagnostically relevant segments (or prototypes) from each pathology report, we compute a single importance score for each segment, by averaging its attention weights across all query positions. This process produces a vector of importance scores for all sections in a report:  $\mathbf{s}^t = [s_1^t, s_2^t, \cdots s_m^t] \in \mathbb{R}^m$, showing, on average, how much the entire report attends to each segment. Mathematically, if $\mathbf{A}_n^t \in \mathbb{R}^{m\times m}$ is the attention matrix for the $n$-th report, we define the importance score $s_{n,j}^t$ of section $j$ as:
\begin{equation}
    s_{n,j}^t = \frac{1}{m}\sum_{i=1}^m A_{n,i,j}^t
\end{equation}
where $A_{n,i,j}^t$ is the attention weight from query position $i$ to key position $j$. The sections are then ranked in descending order of their importance scores. Segments that are more relevant for diagnosis (\eg those describing tumor size or pathology findings) typically receive higher scores. From the post-attention representation $\mathbf{Z}_i^t$, we select embeddings corresponding to the top $N_{\mathcal{T}}$ sections, yielding fixed-length Diagnostic Prototypes: $\mathbf{Z}_{i,\text{proto}}^t \in \mathbb{R}^{ N_{\mathcal{T}} \times d_t}$,  where $N_{\mathcal{T}}$ is the number of prototypes and $d_t$ is the embedding dimension for the $i^{th}$ report. To facilitate multimodal fusion, we align the dimensions of the prototype representations derived from the three modalities. For this reason, we apply a linear transformation $f^t_{\alpha}$ to the text prototype embeddings, resulting in: $\mathbf{Z}^t_{\alpha} = f^t_{\alpha}(\mathbf{Z}_{\text{proto}}^t)\in \mathbb{R}^{d_e}$.

\subsection{Histological prototypes from WSIs}
\label{section: histology prototypes}
\textbf{Feature Extraction: }For each WSI, we first identify and isolate the tissue regions \cite{pocock2022tiatoolbox} to ensure that diagnostically irrelevant background regions are excluded from further analysis. The retained tissue region in each histology image $(h)$, denoted as $X_i$, is divided into non-overlapping patches \cite{lu2021data}  as follows: $X_i = \{x_1^h,x_2^h \ldots x^h_{N_{h,i}}\}$ where $x_j^h \in \mathbb{R}^{H\times W\times3}$ denotes the $j^{th}$ patch, $H$ and $W$ represent the height and width of the patch, respectively and $N_{h,i}$ represents the total number of patches in the $i^{th}$  slide. Each patch is processed by a pre-trained and frozen image encoder $E_I$ to extract visual features: $\textbf{h}_j^h = E^I(x_j^h) \in \mathbb{R}^{d_h}$, where $d_h$ represents the dimensions of the extracted visual feature vector. The set of extracted feature vectors for slide $i$ are collectively represented as $\textbf{H}_i^h = \{\textbf{h}_1^h,\textbf{h}_2^h \ldots \textbf{h}_{N_{h,i}}^h\}$. 

\textbf{Prototype Construction: }To summarize the recurring morphological patterns in WSIs, we cluster patches into $N_\mathcal{H}$ histological prototypes using GMM, inspired by previous methods \cite{song2024morphological,songmultimodal}. Before applying the GMM, we first obtain an initial estimate of the histological prototypes by randomly initializing cluster centers in the feature space. Each prototype mean ($\mu$) is sampled from a Gaussian distribution, ensuring that the cluster centers are spread out. The diagonal covariance matrices  ($\Sigma$) are initialized as identity matrices while the mixture weights  ($\pi$) are uniformly initialized. Mathematically we represent this as: 
\begin{equation}
\mu_c \sim \mathcal{N}(0, 0.1^2 I)
, \quad \Sigma_c = I, \quad \pi_c = \frac{1}{N_\mathcal{H}}
\end{equation}
$\forall c \in \{1, 2, \dots, N_\mathcal{H}\}$ where $N_\mathcal{H}$ denotes the number of prototypes $c$. 
Unlike traditional clustering methods, GMM allows soft assignments, meaning each patch is assigned to multiple prototypes with varying probabilities, enabling a more flexible representation of the morphological diversity found in the tissue structures \cite{song2024morphological}. The likelihood of a patch embedding under the GMM is given by:
\begin{equation}
p(\textbf{h}_{j}^h) = \sum_{c=1}^{N_\mathcal{H}} \pi_{c} \cdot \mathcal{N}(\textbf{h}_{j}^h; \mu_{c}, \Sigma_{c})
\end{equation}
where: $\pi_c$ is the probability of selecting prototype $c$ and $\mathcal{N}(\textbf{h}_{j}^h; \mu_{c}, \Sigma_{c})$ represents the Gaussian density function and models the likelihood of $\textbf{h}_{j}^h$ given prototype $c$.
For an entire WSI, the joint probability over all patches is:
\begin{equation}
p(\textbf{H}^h) = \prod_{j=1}^{N_h} \sum_{c=1}^{N_\mathcal{H}} \pi_{c} \cdot \mathcal{N}(\textbf{h}_{j}^h; \mu_{c}, \Sigma_{c})
\end{equation}
This formulation ensures that the WSI is summarized using a compact set of histological prototypes. The GMM parameters ($\mu, \Sigma, \pi$) are optimized using the Expectation-Maximization (EM) algorithm \cite{dempster1977maximum,kim2022differentiable}. Over successive EM iterations, the prototypes are progressively updated, ultimately converging to meaningful cluster centers. After convergence, the final slide representation is obtained by stacking the estimated mixture parameters:
\begin{equation}
\textbf{Z}_{\text{proto}}^{h} = [\pi_{1}, \mu_{1}, \Sigma_{1}, \dots, \pi_{N_\mathcal{H}}, \mu_{N_\mathcal{H}}, \Sigma_{N_\mathcal{H}}] \in \mathbb{R}^{N_\mathcal{H} \times (1+2d_h)}
\end{equation}
where: $\pi_c$ quantifies the prevalence of each prototype in the slide, $\mu_c $ represents the average morphological characteristics of each prototype and $\sum_c$ captures the variation within the patches belonging to each prototype. This prototype-based embedding significantly reduces the dimensionality of WSI patch features whilst still preserving key morphological characteristics. To align the dimensionality of histology prototypes with those from other modalities, we apply a linear transformation $f^h_{\alpha}$ to the prototype embeddings, yielding : $\mathbf{Z}^h_{\alpha} = f^h_{\alpha}(\mathbf{Z}_{\text{proto}}^h)\in \mathbb{R}^{d_e}$.

\subsection{Pathway Prototypes from Genomic Data}
\label{section: pathway prototypes}
\textbf{Prototype Construction:}  The transcriptomic profile for each sample can be represented by a gene-expression vector $\mathbf{x}^p \in \mathbb{R}^{N_G}$, where $N_G$ is the total number of genes. We aim to group these genes $(p)$ into $N_{\mathcal{P}}$ pre-defined biological pathways, each acting as a prototype. For each pathway $i$ we define a binary mask $\mathbf{B}_i \in \{0,1\}^{N_G}$, where each element indicates whether a gene is included in a given pathway ($1$) or not ($0$). 
To construct the pathway prototypes, we perform an element-wise product of the gene expression vector $(\mathbf{x}^p)$ with the binary pathway vectors ($\mathbf{B}$) \cite{songmultimodal}. This operation generates a pathway-specific representation:
\begin{equation}
    \mathbf{Z}_{\text{proto},i}^p = \mathbf{x}^p \odot \mathbf{B}_i, \quad \forall i \in [1, N_{\mathcal{P}}]
\end{equation}
where $N_{\mathcal{P}}$ denotes the number of pathways. The resulting vector 
$ \mathbf{Z}_{\text{proto},i}^p$ is then reduced to remove any zero entries, yielding a dense but variable-length representation (since different pathways involve different gene counts).
To ensure a fixed size embedding for each pathway, we employ self-normalizing neural networks (SNNs) \cite{klambauer2017self,jaume2024modeling} :
\begin{equation}
    \mathbf{Z}_{i, \alpha}^p = f_{\alpha,i}^p(\mathbf{Z}_{\text{proto}, i}^p )\in \mathbb{R}^{d_e} , \forall i\in [1,N_{\mathcal{P}}]
\end{equation}
where $f_{\alpha,i}^p$ is a trainable network dedicated to pathway $i$. By treating each pathway as a “prototype” we are able to preserve biologically relevant groupings of genes while ensuring that the final representation for each pathway shares the same dimensionality with itself and aligns with the other modalities.

\subsection{Multimodal Fusion}
\label{section: fusion}
We aim to design a multimodal fusion model that extends the standard Transformer attention mechanism \cite{vaswani2017attention,dosovitskiy2020image} to capture both intra-modal and cross-modal interactions. Specifically, we seek to learn interactions among: diagnostic prototypes from pathology reports ($t$) represented as $\mathbf{Z}^{t}_{\alpha} \in \mathbb{R}^{N_\mathcal{T}}$, histological prototypes from histology images ($h$) represented as $\mathbf{Z}^{h}_{\alpha} \in \mathbb{R}^{N_\mathcal{H}}$ and pathway prototypes from transcriptomic data ($p$) represented as $\mathbf{Z}^{p}_{\alpha} \in \mathbb{R}^{N_\mathcal{P}}$. Each modality-specific embedding consists of $N_\mathcal{M}$ prototypes, where $\mathcal{M}\in \{\mathcal{P,H,T}\}$. To allow the model to dynamically enhance the prototype representations, we also append a learnable and randomly initialized embedding $\mathbf{e}_r \in \mathbb{R}^{d_r}$ to the prototypes \cite{songmultimodal,jaegle2021perceiver,liang2022high}. The modified embeddings are then represented as $\mathbf{Z}_{\alpha}^{h,p,t} \in \mathbb{R}^{d = d_e+d_r}$.

To integrate the three modalities, we concatenate the modality-specific prototypes into a single multimodal embedding:  $\mathbf{Z}^{mm} = [\mathbf{Z}^{p}_{\alpha} \mathbin{\|} \mathbf{Z}^h_{\alpha}  \mathbin{\|} \mathbf{Z}^t_{\alpha}]\in \mathbb{R}^{(N_\mathcal{P}+N_{\mathcal{H}}+N_{\mathcal{T}}) \times d}$, where $d$ is the feature dimension. Following the standard Transformer architecture, we introduce three learnable projection matrices, $\mathbf{W}_Q , \mathbf{W}_K, \mathbf{W}_V \in \mathbb{R}^{d \times d}$ to map the multimodal embedding $\mathbf{Z}^{mm}$ into queries, keys and values:
\begin{equation}
\mathbf{Q} = \mathbf{W}_Q \cdot {\mathbf{Z}}^{mm},\quad\mathbf{K} = \mathbf{W}_K \cdot {\mathbf{Z}}^{mm},\quad\mathbf{V} = \mathbf{W}_V\cdot {\mathbf{Z}}^{mm} 
\end{equation}
However, rather than treating $\mathbf{Z}^{mm}$ as a unified sequence, we decompose it into modality-specific components and compute attention for all possible pairwise interactions. Concretely, we split each $\mathbf{Q}$, $\mathbf{K}$, $\mathbf{V}$ into three modality-specific blocks:
\begin{equation}
\begin{split}
\mathbf{Q} = \begin{bmatrix} \mathbf{Q}^p \parallel \mathbf{Q}^h \parallel \mathbf{Q}^t \end{bmatrix}, \\
\mathbf{K} = \begin{bmatrix} \mathbf{K}^p \parallel \mathbf{K}^h \parallel \mathbf{K}^t \end{bmatrix}, \\
\mathbf{V} = \begin{bmatrix} \mathbf{V}^p \parallel \mathbf{V}^h \parallel \mathbf{V}^t \end{bmatrix},
\end{split}
\end{equation}
where $\mathbf{Q}^p, \mathbf{K}^p, \mathbf{Q}^p \in \mathbb{R}^{N_\mathcal{P} \times d}$ for pathways (and likewise for histology and text). 

The approach results in a total of nine attention mechanisms categorized into $(1)$ three self-attention (intra-modal) interactions: $\mathbf{A}^{p \rightarrow p}$ (pathways attending to pathways) , $\mathbf{A}^{h \rightarrow h}$(histology attending to histology) and $\mathbf{A}^{t \rightarrow t}$ (text attending to text) $(2)$ six cross-attention (inter-modal) interactions:  $\mathbf{A}^{p \rightarrow h},  \mathbf{A}^{p \rightarrow t}$ (pathways attending to histology and text), $ \mathbf{A}^{h \rightarrow p},  \mathbf{A}^{h \rightarrow t}$ (histology attending to pathways and text) and $\mathbf{A}^{t \rightarrow p}, \mathbf{A}^{t \rightarrow h}$ (text attending to pathways and histology). Each attention component is computed as: $\mathbf{A}^{m \rightarrow n} = \mathbf{Q}^{m} (\mathbf{K}^{n})^{\top} $ where $\mathbf{A}^{m \rightarrow n}$ represents the attention logits from modality $m$ to modality $n$.

After computing all intra- and cross-modal attention scores, we concatenate the sub-blocks within each modality and apply a row-wise softmax operation, producing a single probability distribution over key tokens for each query modality. This ensures that each modality attends to the most relevant information from both itself and the other modalities. Once the attention scores are computed, they are multiplied by the corresponding value representations from all modalities. The final modality-specific embeddings after fusion are:

\begin{equation}
\begin{split}
\begin{pmatrix}
{\mathbf{Z}}^{p}_{\text{fusion}} \\[6pt]
{\mathbf{Z}}^{h}_{\text{fusion}} \\[6pt]
{\mathbf{Z}}^{t}_{\text{fusion}}
\end{pmatrix}
= 
\sigma\left[
  \frac{1}{\sqrt{d}}
  \begin{pmatrix}
    \mathbf{A}^{p \rightarrow p} & \mathbf{A}^{p \rightarrow h} & \mathbf{A}^{p \rightarrow t} \\[6pt]
    \mathbf{A}^{h \rightarrow p} & \mathbf{A}^{h \rightarrow h} & \mathbf{A}^{h \rightarrow t} \\[6pt]
    \mathbf{A}^{t \rightarrow p} & \mathbf{A}^{t \rightarrow h} & \mathbf{A}^{t \rightarrow t}
  \end{pmatrix}
\right]
\begin{pmatrix}
\mathbf{V}^{p} \\[3pt]
\mathbf{V}^{h} \\[3pt]
\mathbf{V}^{t}
\end{pmatrix} \\
= 
\sigma\left[
  \frac{1}{\sqrt{d}}
  \begin{pmatrix}
    \mathbf{Q}^{p} (\mathbf{K}^{p})^{\top} & \mathbf{Q}^{p} (\mathbf{K}^{h})^{\top} & \mathbf{Q}^{p} (\mathbf{K}^{t})^{\top} \\[6pt]
    \mathbf{Q}^{h} (\mathbf{K}^{p})^{\top} & \mathbf{Q}^{h} (\mathbf{K}^{h})^{\top} & \mathbf{Q}^{h} (\mathbf{K}^{t})^{\top} \\[6pt]
    \mathbf{Q}^{t} (\mathbf{K}^{p})^{\top} & \mathbf{Q}^{t} (\mathbf{K}^{h})^{\top} & \mathbf{Q}^{t} (\mathbf{K}^{t})^{\top}
  \end{pmatrix}
\right]
\begin{pmatrix}
\mathbf{V}^{p} \\[3pt]
\mathbf{V}^{h} \\[3pt]
\mathbf{V}^{t}
\end{pmatrix}
\label{eq:full-block-attn}
\end{split}
\end{equation}
where $\sigma[\cdot]$ is the row-wise softmax operation. Finally, the full multimodal representation is obtained by concatenating the post-fusion modality-specific embeddings: 
\begin{equation}
    \mathbf{Z}_{\text{fusion}}^{mm} = [{\mathbf{Z}}^{p}_{\text{fusion}}\|{\mathbf{Z}}^{h}_{\text{fusion}}\|{\mathbf{Z}}^{t}_{\text{fusion}}]
\end{equation}

\subsection{Survival Prediction}
\label{section: post processing}
After the multimodal fusion step, the resulting embeddings are processed by a series of multi-layer perceptrons (MLPs) for further transformation. Within each modality, these embeddings undergo layer normalization and are subsequently averaged to create compact, modality-specific representations \cite{songmultimodal}. 
\begin{equation}
risk = f_{risk}
\left[
\frac{1}{N_{\mathcal{X}}} \sum_{i=1}^{N_{\mathcal{X}}} f_{\beta}( \mathbf{Z}^{x}_{i,\text{fusion}} )
\right]_{\mathcal{X} \in \{ \mathcal{P}, \mathcal{H}, \mathcal{T} \}}
\end{equation}
The final patient-level representation is obtained by concatenating these averaged modality-specific embeddings, which are then passed through a linear layer $f_{\text{risk}}$ to generate a patient-level prediction. 

\section{Datasets and Implementation}
\subsection{Datasets}
We evaluate our proposed method using publicly available datasets from The Cancer Genome Atlas (TCGA) across six types of cancer \cite{TCGA}: Bladder urothelial carcinoma (BLCA) (n = $328$), Lung adenocarcinoma (LUAD) (n = $391$), Kidney renal clear cell carcinoma (KIRC) (n = $328$ ), Stomach adenocarcinoma (STAD) (n = $253$), Colon and Rectum adenocarcinoma (CRC) (n = $269$) and Head and Neck Squamous Cell Carcinoma (HNSC) (n= $385$). 

Hematoxylin and Eosin (H\&E) stained WSIs for all patients were obtained from the National Cancer Institute (NCI) Genomic Data Commons (GDC) \cite{TCGA}. The corresponding pathology reports were obtained from Kefeli \etal.\cite{kefeli2024tcga}, who curated 9,523 machine-readable reports from TCGA by converting PDF documents into text format. Bulk RNA sequencing data for TCGA cohorts was accessed from the UCSC Xena database \cite{goldman2020visualizing}, measured using the Illumina HiSeq 2000 system. To stabilize variance and mitigate the impact of extreme values, they applied $log_2 (x+1)$ transformation and RSEM normalization \cite{li2011rsem}. To structure the data into pathways, we utilized 50 Hallmark gene sets from the Molecular Signatures Database (MSigDB) \cite{subramanian2005gene,liberzon2015molecular}, covering approximately 4000 unique genes in total. Further implementation details can be found in Supplementary Material, \textbf{Implementation Details}.

\subsection{Baselines}
\subsubsection{Unimodal Baselines}
The following unimodal baselines were separately implemented for each data modality:

\textbf{Text:} We employ a unimodal text-only variant of the proposed method (with prototypes), $\text{PS3}_{\text{t}}$. Additionally, we use an Attention-based MIL \cite{ilse2018attention} on the pathology reports. \textbf{Gene:} We use a unimodal transcriptomics-only variant of the proposed method (with prototypes), $\text{PS3}_{\text{p}}$. Additionally, we use a 2-layer MLP as a non-prototype baseline \cite{songmultimodal,jaume2024modeling}. \textbf{Histology:} We utilize TransMIL \cite{shao2021transmil}, $\text{R}^{2}\text{-TMIL}$ \cite{tang2024feature} and CLAM \cite{lu2021data,yang2024foundation} as our histology baselines. We also implement a unimodal histology-only variant of the proposed method (with prototypes), $\text{PS3}_{\text{WSI}}$. 

\subsubsection{Multimodal Baselines}
We implement MOTCat \cite{xu2023multimodal}, MCAT \cite{chen2021multimodal}, SurvPath \cite{jaume2024modeling}, SurvivMIL \cite{naidoo2024survivmil}, CMTA \cite{zhou2023cross}, FSM \cite{zheng2024graph} and both variants of MMP \cite{songmultimodal} 
- $\text{MMP}_{\text{Trans}}$ and 
$\text{MMP}_{\text{OT}}$ — as our multimodal baselines. Additional details can be found in Supplementary Material, \textbf{Multimodal Baselines}.


\section{Results}

\begin{table*}[ht]
\small
    \centering
    \begin{tabular}{p{0.01cm}l|c|c|c|c|c|c|c}
        \toprule
         & \textbf{Model} & \textbf{BLCA} & \textbf{LUAD} & \textbf{KIRC} & \textbf{STAD} & \textbf{CRC} & \textbf{HNSC} & \textbf{Avg ($\uparrow$)} \\
        \midrule
        & Clinical & $0.557 \pm 0.062 $ & $0.494 \pm 0.093$ & $0.723 \pm 0.044$ & $0.583 \pm  0.051$ & $0.496 \pm 0.099$ & $0.516 \pm 0.090 $ & $0.561$ \\
        \midrule
        \multirow{2}{*}{\rotatebox[origin=c]{90}{\textbf{Gene}}} 
            & Gene exp \cite{jaume2024modeling} & $0.656 \pm 0.047$ & $0.508 \pm 0.064$ & $0.764 \pm 0.042$ & $0.578 \pm 0.080$ & $0.678 \pm 0.069$ & $0.595 \pm 0.076$ & $0.630$ \\
            & $\text{PS3}_{\text{p}}$ & $0.643 \pm 0.062$ & $0.578 \pm 0.068$ & $0.765 \pm 0.040$ & $0.614 \pm 0.059$ & $0.710 \pm 0.063$ & $0.601 \pm 0.060$ & $0.652$ \\
        \midrule
        \multirow{2}{*}{\rotatebox[origin=c]{90}{\textbf{Text}}} 
            & Text ABMIL \cite{ilse2018attention} & $0.522 \pm 0.069$ & $0.560 \pm 0.053$ & $0.598 \pm 0.079$ & $0.594 \pm 0.067$ & $0.772 \pm 0.137$ & $0.507 \pm 0.067$ & $0.592$ \\
            & $\text{PS3}_{\text{t}}$ & $0.629 \pm 0.065$ & $0.536 \pm 0.087$ & $0.623 \pm 0.075$ & $0.548 \pm 0.078$ & $0.805 \pm 0.130$ & $0.550 \pm 0.080$ & $0.615$ \\
        \midrule
        \multirow{4}{*}{\rotatebox[origin=c]{90}{\textbf{Histology}}} 
            & CLAM \cite{lu2021data} & $0.562 \pm 0.100$ & $0.606 \pm 0.104$ & $0.684 \pm 0.055$ & $0.510 \pm 0.078$ & $0.680 \pm 0.127$ & $0.560 \pm 0.075$ & $0.600$ \\
            & TransMIL \cite{shao2021transmil} &$0.585 \pm 0.049$ & $0.594 \pm 0.096$ & $0.736 \pm 0.085$ & $0.558 \pm 0.035$ & $0.671 \pm 0.120$ & $0.594 \pm 0.070$ & $0.623$ \\
            & RRTMIL \cite{tang2024feature} &$0.550 \pm 0.065$ & $0.557 \pm 0.091$ & $0.704 \pm 0.128$ & $0.603 \pm 0.088$ & $0.575 \pm 0.035$ & $0.556 \pm 0.056$ & $0.591$ \\
            & $\text{PS3}_{\text{WSI}}$ &$0.539 \pm 0.033$ & $0.597 \pm 0.071$ & $0.759 \pm 0.102$ & $0.542 \pm 0.085$ & $0.586 \pm 0.197$ & $0.502\pm 0.053$ & $0.588$ \\
        \midrule
        \multirow{7}{*}{\rotatebox[origin=c]{90}{\textbf{Multimodal}}} 
            & SurvivMIL \cite{naidoo2024survivmil} &$0.484 \pm 0.050$ & $0.568 \pm 0.048$ & $0.654 \pm 0.101$ & $0.492 \pm 0.068$ & $0.663 \pm 0.057$ & $0.552 \pm 0.086$ & $0.569$ \\
            & SurvPath \cite{jaume2024modeling} &$0.613 \pm 0.036$ & $0.567 \pm 0.055$ & $0.761 \pm 0.054$ & $0.608 \pm 0.048$ & $0.640 \pm 0.054$ & $0.536 \pm 0.055$ & $0.621$ \\
            & MOTCat \cite{xu2023multimodal} &$0.636 \pm 0.057$ & $0.533 \pm 0.039$ & $0.766 \pm 0.049$ & $0.553 \pm 0.082$ & $0.677 \pm 0.067$ & $0.586 \pm 0.044$ & $0.625$ \\
            & MCAT \cite{chen2021multimodal} & $0.636 \pm 0.068$ & $0.512 \pm 0.040$ & $0.762 \pm 0.030$ & $0.572 \pm 0.074$ & $0.661 \pm 0.101$ & $0.578 \pm 0.064$ & $0.620$ \\
            & CMTA \cite{zhou2023cross} & $0.637 \pm 0.067$ & $0.565 \pm 0.045$ & $0.741 \pm 0.044$ & $0.578 \pm 0.065$ & $0.659 \pm 0.058$ & $0.579 \pm 0.030$ & $0.627$ \\
            & FSM \cite{zheng2024graph} & $0.642 \pm 0.050$ & $0.565 \pm 0.082$ & $\textbf{0.776} \pm 0.048$ & $0.609 \pm 0.068$ & $0.663 \pm 0.059$ & $0.577 \pm 0.064$ & $0.639$ \\
            & $\text{MMP}_{\text{Trans}}$ \cite{songmultimodal} &$0.641 \pm 0.053$ & $0.606 \pm 0.068$ & $\textbf{0.776} \pm 0.059$ & $0.639 \pm 0.063$ & $0.689 \pm 0.078$ & $0.566 \pm 0.075$ & $0.653$ \\
            & $\text{MMP}_{\text{OT}}$ \cite{songmultimodal} &$0.645 \pm 0.030$ & $0.617 \pm 0.058$ & $0.774 \pm 0.026$ & $\textbf{0.660} \pm 0.073$ & $0.689 \pm 0.074$ & $0.545 \pm 0.041$ & $0.655$ \\
            \midrule
            & \textbf{PS3}& $\textbf{0.684} \pm 0.026$ & $\textbf{0.640} \pm 0.093$ & $\textbf{0.776} \pm 0.061$ & $0.638 \pm 0.045$ & $\textbf{0.826} \pm 0.101$ & $\textbf{0.627} \pm 0.066$ & $\textbf{0.699}$ \\
        \bottomrule
    \end{tabular}
    \caption{Survival prediction results comparing the proposed method, PS3, with multimodal and unimodal baselines for disease-specific survival prediction using the C-Index. Pathology Language and Image Pre-Training (PLIP) \cite{huang2023visual} is used as a feature encoder across all methods. Performance is evaluated over five runs, with standard deviation reported. The best-performing results are highlighted in bold.}
    \label{table:c index results}
\end{table*}

\subsection{Survival Prediction Results}
Table.\ref{table:c index results} presents the C-Index results for PS3 and all baseline models in predicting disease-specific survival. PS3 demonstrates the highest overall (average) performance, outperforming the next-best multimodal and unimodal methods with percentage improvements of $6.72\%$ and $7.21\%$, respectively. In addition, it achieves among the highest performance in 5 of the 6 cancer types evaluated. We summarize the key findings below. Additional results can be found in Supplementary Material, \textbf{Kaplan-Meier Analysis} and \textbf{Attention Visualization}.

\subsubsection{Comparison with Clinical Baseline}
Clinical variables such as age, sex and histologic grade have been identified as important prognostic factors for survival prediction \cite{tas2013age,vlenterie2015age,joosse2013sex,sun2006histologic}. In Table.\ref{table:c index results} our analysis demonstrates that all subsequent methods achieve superior performance (by a minimum of $1.4\%$) compared to the clinical baseline, which relies solely on these variables. This highlights the benefits of integrating additional data modalities such as histology images, transcriptomic data and pathology reports. Additional details can be sound in Supplementary Material, \textbf{Clinical Baselines}.

\begin{table}[ht]
\small
    \centering
    \begin{tabular}{c|c|c}
    \toprule
    \textbf{Ablation} & \textbf{Model} & \textbf{Avg.} \\ 
    \midrule
     & \textbf{PS3}& \textbf{0.699} \\ 
    \midrule
    Text Proto $N_{\mathcal{T}}$ & Avg $\Rightarrow$ p90 & $0.691$ ($-1.14\%$) \\ 
    \midrule
    Encoder $E_I, E_T$ & PLIP $\Rightarrow$ QUILT-Net & $0.676$ ($-3.29\%$) \\ 
    \midrule
    \multirow{3}{*}{Modalities} & $p,h,t$ $\Rightarrow$ $h,t$  & $0.644 $ ($-7.87\%$) \\ 
    & $p,h,t$ $\Rightarrow$ $p,t$  & $0.687$ ($-1.72\%$) \\ 
    & $p,h,t$ $\Rightarrow$ $p,h$  & $0.653$ ($-6.58\%$) \\ 
    \midrule
    \multirow{2}{*}{Fusion Method} & Full $\Rightarrow$ Late & $0.669$ ($-4.29\%$) \\ 
    & Full $\Rightarrow$ Hierarchical & $0.660$ ($-5.58\%$) \\ 
    \midrule
    {Embeddings $\mathbf{e}_r$} & Random $\Rightarrow$ None & $0.688$ ($-1.57\%$) \\ 
    \midrule
    MLP $f_{\beta}$& Multiple $\Rightarrow$ Single & $0.690$ ($-1.29\%$) \\
    \bottomrule
    \end{tabular}
    \caption{Ablation study analyzing the impact of modifying individual model components on the C-Index, with results averaged across six cohorts.}
\label{table:Ablation}
\end{table}

\subsubsection{Comparison of Histology, Gene and Text-Based Methods}
Table.\ref{table:c index results} also demonstrates that gene-based methods (with C-Index results $0.63-0.65$) outperform other unimodal baselines, highlighting the strong prognostic value of transcriptomics data for survival prediction. In contrast, histology- (with C-Index results $0.58-0.62$) and text-based methods (with C-Index results $0.59-0.61$) achieve similar performance, likely due to their shared reliance on histopathological features. While WSIs provide direct morphological insights and pathology reports summarize these observations, both modalities may be more affected by feature extraction challenges and variability in reporting, making them less discriminative than genomic features.

\subsubsection{Impact of Prototypes in Unimodal and Multimodal Approaches}
In unimodal settings, prototype-based methods outperform non-prototype approaches for gene and text data. However, for histology, non-prototype methods achieve higher performance, as seen in Table.\ref{table:c index results}, where TransMIL, CLAM and $\text{R}^{2}\text{-TMIL}$ outperform $\text{PS3}_{\text{WSI}}$. This decline can be attributed to the compression effect of histological prototyping. Reducing thousands of image patches to a compact set of prototypes reduces data dimensionality and complexity but can lead to the loss of fine-grained histological details.

However, prototyping can enhance multimodal learning by facilitating better integration across modalities. While MIL-based methods perform well in unimodal settings, they tend to struggle with multimodal integration. In contrast, prototyping-based approaches significantly outperform MIL-based methods when combining histology with transcriptomics and/or text. As shown in Table.\ref{table:c index results},  $\text{PS3}$, $\text{MMP}_{\text{Trans}}$ and $\text{MMP}_{\text{OT}}$ outperform SurvPath, MCAT, CMTA and MOTCat, demonstrating the benefits of prototyping for multimodal fusion. A similar trend is observed in the histology-text configuration where the  histology $+$ text method $(h,t)$ ablation study in Table.\ref{table:Ablation} outperforms MIL-based method, SurvivMIL by $13.18\%$. Data scale disparities across modalities can lead to modality dominance and imbalance, hindering effective multimodal integration. Prototyping mitigates this by normalizing data scale while preserving essential information, ensuring more balanced multimodal integration where each modality contributes meaningfully to survival prediction.


\subsubsection{ Two-Modal vs. Three-Modal Approaches} 
\label{sec:results_twovsthree}
As shown in Table.\ref{table:c index results}, integrating all three modalities outperforms both dual-modality and unimodal approaches, achieving $6.72\%$ and $7.21\%$ improvement compared to the next-best multimodal (with C-Index $0.655$) and unimodal (with C-Index $0.652$) methods, respectively. This underscores the value of incorporating pathology reports into the workflow. Further supporting this, Table.\ref{table:Ablation} presents an ablation study (Impact of Modality Combinations), which evaluates our model’s performance across different two-modality combinations. The results show that the proposed three-modal approach, PS3 $(h,p,t-0.699)$ achieves $8.54\%$ higher performance than Histology + Text ($h,t-0.644$), $1.75\%$ higher performance than Pathways + Text ($p,t-0.687$) and $7.04\%$ higher performance than Pathways + Histology ($p,h-0.653$). Notably, Pathways + Text outperforms both other dual-modality combinations, highlighting the prognostic value of pathology reports. These results highlight the advantage of multimodal integration in improving patient prognostication.

\subsection{Ablation Study}
We conducted extensive ablation studies in Table.\ref{table:Ablation} to evaluate the impact of various design choices on multimodal learning performance. Below, we summarize our key findings:
$(1)$ \textbf{Number of Diagnostic Text Prototypes :} Instead of setting the number of text prototypes $N_{\mathcal{T}}$ to the average length of text segments in the training dataset, we experimented with setting it to the 90th percentile (p90). However, this introduced additional noise, which reduced the overall performance. $(2)$ \textbf{Feature Encoder Performance:} We compared different vision-language feature encoders and discovered that PLIP-based \cite{huang2023visual} features outperformed those derived from QUILT-Net \cite{ikezogwo2024quilt}. $(3)$ \textbf{Impact of Modality Combinations:} As explained previously in Section.\ref{sec:results_twovsthree} the proposed model which incorporates all three modalities $(h,p,t)$ outperforms all two-modality combinations, highlighting the importance of integrating multiple complementary data sources. $(4)$ \textbf{Fusion Methods Comparison:} We evaluated two different fusion strategies: 
\textit{Late Fusion}: This approach only applies attention within each modality, without capturing any cross-modal interactions. Modalities are only merged at the final stage before prediction. \textit{Hierarchical Fusion}: This approach employs a two-step process. First, it models cross- and self-modal interactions between histology and text. Then, the fused (histology $+$ text) representation undergoes additional self and cross-attention with transcriptomics. We note that the proposed method outperforms both aforementioned fusion strategies. $(5)$ \textbf{Effect of Learnable Embeddings:} We experimented without adding the randomly initialized embeddings, $\mathbf{e}_r$ to the feature representations of each modality. Our results indicate that these embeddings improve model performance by allowing the model to learn richer feature representations.
$(6)$ \textbf{ MLPs:} Instead of separate MLPs for the prototypes we experiment with implementing a single shared MLP for all prototypes across different modalities.

\section{Conclusions}
Pathology reports contain critical diagnostic and prognostic insights, yet remain under-utilized in computational survival prediction models. While transcriptomic and histological data capture molecular and spatial characteristics, pathology reports offer expert-driven interpretations that complement these modalities. However, existing approaches often overlook this valuable resource in multimodal survival analysis. To address this, we proposed a prototype-based multimodal fusion framework that integrates cancer aggressiveness-related signals from WSIs, pathology reports and transcriptomic data for improved survival prediction. By transforming each modality into structured representations -- histological prototypes for WSIs, diagnostic prototypes for pathology reports and pathway prototypes for transcriptomic data -- we mitigated modality imbalance and improved predictive performance. The PS3 multimodal transformer effectively modeled cross-modal interactions across all three modalities, leading to superior predictive performance. We evaluated our proposed model across six cancer types, outperforming both unimodal and multimodal baselines.

Future work can further enhance this approach by incorporating additional multi-omics modalities, radiology images and patient metadata, to provide a more comprehensive view of patient characteristics. By continuing to refine multimodal fusion strategies, we can advance precision oncology and improve patient outcomes.
\section*{Acknowledgments}
The first author would like to acknowledge support from AstraZeneca, UK and the Department of Computer Science at the University of Warwick, UK.

{
    \small
    \bibliographystyle{ieeenat_fullname}
    \bibliography{main}
}

\newpage
\section*{Supplementary Information}
\setcounter{section}{0} 
\renewcommand{\thesection}{S\arabic{section}} 

\section{Implementation Details}
\label{Supplementary:Implementation Details}
We employ Pathology Language and Image Pre-Training (PLIP) \cite{huang2023visual} as our encoder to extract both image and text features from the WSIs and their corresponding pathology reports. As part of our ablation study, we also experiment with QUILT-Net \cite{ikezogwo2024quilt} as an alternative feature extractor. Since our approach requires extracting both image and text embeddings, we are inherently constrained to vision-language models for feature extraction.

Both PLIP and QUILT-Net are vision-language models \cite{bilal2025foundation} that fine-tune a pretrained contrastive language-image pretraining (CLIP) model \cite{radford2021learning}. PLIP is trained on OpenPath, a dataset consisting of approximately 200,000 paired pathology image-text pairs, curated from publicly available sources such as medical Twitter \cite{huang2023visual}. Similarly, QUILT-Net is trained on Quilt-1M, a dataset consisting of 1 million pathology image and text samples, sourced from educational histopathology videos along with other publicly available resources \cite{ikezogwo2024quilt}.

We train the models to predict disease-specific survival (DSS) \cite{liu2018integrated}, employing 5-fold site-stratified cross-validation \cite{howard2021impact}, a widely used approach in the literature. Model performance is evaluated using the concordance index (C-Index)  \cite{harrell1996multivariable}, which measures how accurately the model's predicted risks align with actual patient survival outcomes. All models were trained for 50 epochs, utilizing visual and/or text features extracted using the PLIP feature encoder \cite{huang2023visual}. The training process employed a learning rate of $1 \times 10^{-4}$, a weight decay of $1 \times {10}^{-5}$, a cosine learning rate scheduler, and the AdamW optimizer. For MIL-based methods, during training, 4,096 patches were randomly sampled for each WSI. During inference, the entire WSI was processed to generate predictions. MIL-based models were trained using the negative log-likelihood (NLL) loss \cite{zadeh2020bias} with a batch size of 1, while prototype-based models were optimized with Cox loss \cite{cox1972regression} and a batch size of 64. For prototype-based methods, we set the number of histological prototypes to 16, pathway prototypes to 50 and the number of diagnostic prototypes is set to the average length of reports in the training dataset.

\section{Multimodal Baselines}
Among the Multimodal Baselines, MOTCat \cite{xu2023multimodal}, MCAT \cite{chen2021multimodal}, SurvPath \cite{jaume2024modeling} and $\text{MMP}_{\text{Trans}}$ \cite{songmultimodal} utilize transformer-based architectures. With the exception of SurvivMIL \cite{naidoo2024survivmil}, all aforementioned models integrate histology images with genomic data for survival prediction. In contrast, SurvivMIL incorporates histology images and pathology reports, making it the only multimodal baseline that integrates text data. Additionally, all pathology-genomics baselines utilize genomic prototypes by grouping genes into either functional categories \cite{liberzon2015molecular,chen2021multimodal,xu2023multimodal,zhou2023cross} or biological pathways \cite{elmarakeby2021biologically,songmultimodal,jaume2024modeling,reimand2019pathway}. However, only the two MMP variants incorporate both histology and pathway prototypes.

\section{Clinical Baselines}
We conduct both univariate and multivariate Cox regression analyses using clinical variables such as age, sex, and histologic grade. The results in Table.\ref{table:clinical baselines} highlight our method's performance in comparison to individual clinical variables as well as their combined effect. 

\begin{table*}[ht]
\small
    \centering
    \begin{tabular}{l|c|c|c|c|c|c|c}
        \toprule
         \textbf{Model} & \textbf{BLCA} & \textbf{LUAD} & \textbf{KIRC} & \textbf{STAD} & \textbf{CRC} & \textbf{HNSC} & \textbf{Avg ($\uparrow$)} \\ \midrule
         Age & $0.562 \pm 0.064$ & $0.485 \pm 0.093$ & $0.558 \pm 0.075$ & $0.542 \pm 0.096$ & $0.452 \pm 0.153$ & $0.490 \pm 0.030$ & $0.523$ \\
         Sex & $0.484 \pm 0.053$ & $0.533 \pm 0.050$ & $0.521 \pm 0.051$ & $0.554 \pm 0.055$ & $0.556 \pm 0.065$ & $0.488 \pm 0.046$ & $0.515$ \\
         Grade & $0.512 \pm 0.011$ & n/a & $0.731 \pm 0.052$ & $0.560 \pm 0.039$ & n/a & $0.544 \pm 0.059$ & n/a \\ 
         All & $0.557 \pm 0.062 $ & $0.494 \pm 0.093$ & $0.723 \pm 0.044$ & $0.583 \pm  0.051$ & $0.496 \pm 0.099$ & $0.516 \pm 0.090 $ & $0.561$ \\
         \textbf{PS3} & $0.684 \pm 0.026$ & $0.662 \pm 0.102$ & $0.774 \pm 0.067$ & $0.638 \pm 0.045$ & $0.826 \pm 0.101$ & $0.627 \pm 0.066$ & $0.702$ \\
        \bottomrule
    \end{tabular}
    \caption{Survival Prediction Using Clinical Variables: The variables include age, sex, and histologic grade, collectively referred to as "All."}
    \label{table:clinical baselines}
\end{table*}

\section{Attention Visualization}

\begin{figure*}[ht]
\includegraphics[width=\linewidth]{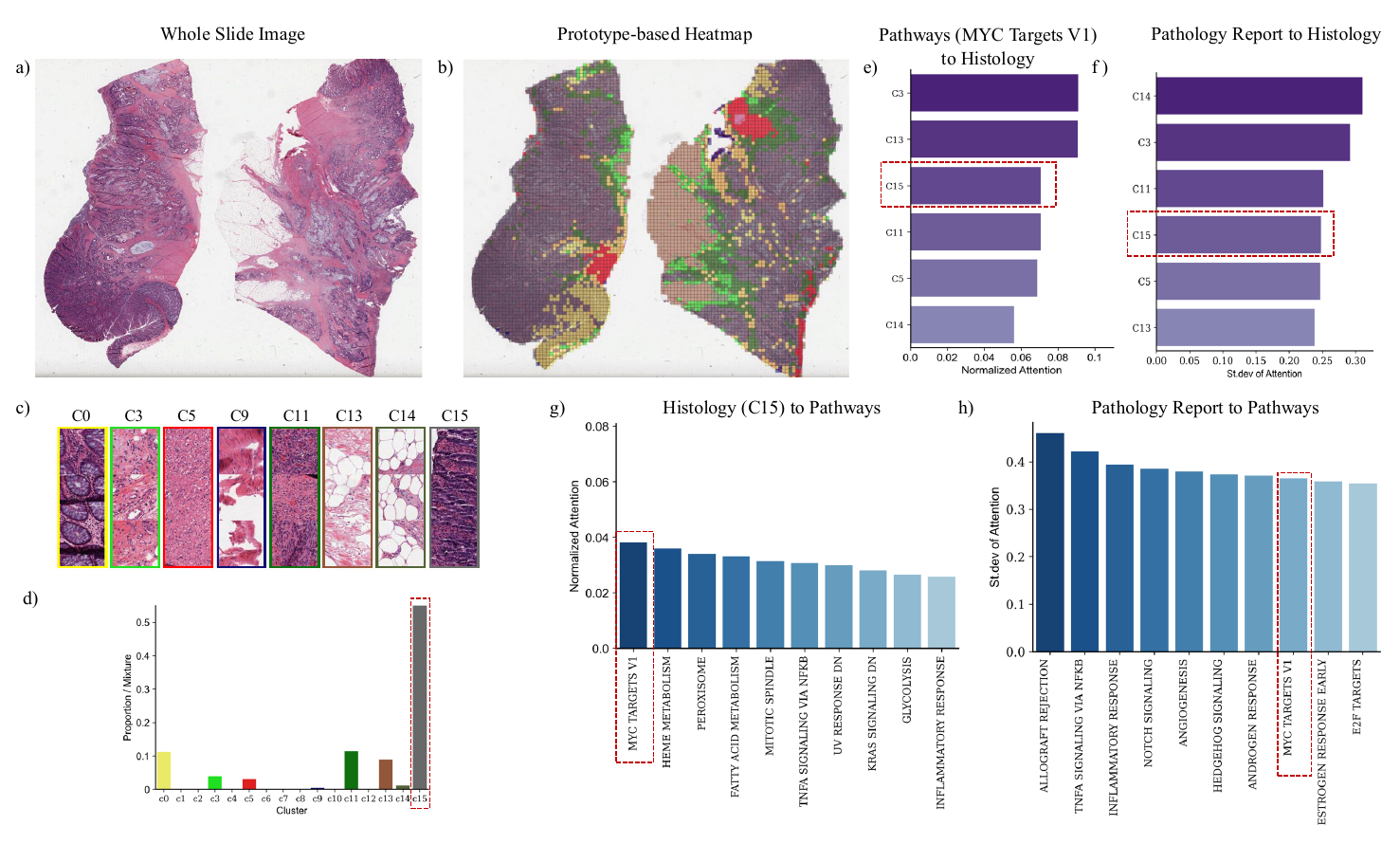}
\caption{(a) A WSI for a CRC patient. (b) Prototype-based heatmap showing
the closest morphological prototype for each patch in the WSI. (c) Top three representative patches for the most significant prototypes. (d)  Proportion of each prototype in the WSI. (e) Top six histological prototypes highly attended by the pathway MYC Targets V1. (f) Top six histological prototypes highly attended by the pathology report. (g) Top ten pathways highly attended by C15 (tumor prototype). (h) Top ten pathways highly attended by the pathology report.}
\label{Figure:Attention}
\end{figure*}

We visualize the histological prototypes created from the WSI and the cross attention between the different modalities \cite{song2024morphological,songmultimodal}.
Each WSI is represented by a compact set of 16 histological prototypes.  Figure.\ref{Figure:Attention}.a represents a TCGA-CRC WSI while Figure.\ref{Figure:Attention}.b displays a heatmap showing the spatial distribution of patches corresponding to each prototype. Figure.\ref{Figure:Attention}.d illustrates the proportion of patches assigned to each prototype ($c$), while Figure.\ref{Figure:Attention}.c highlights representative patches from the most significant prototypes - those with a substantial number of assigned patches. The histological prototypes have been annotated by a pathologist to provide meaningful interpretations. Prototype 0 is associated with normal colon crypts, and 3 captures fibrous connective tissue. Smooth muscle is represented by prototypes 5 and 9, whereas prototype 13 includes both fibrous and adipose tissue. Prototype 14 corresponds specifically to adipose tissue. Lastly, Prototype 15 represents tumor regions, while 11 corresponds to tumor stroma. 

We model cross-modal attention across histology, pathways, and text, capturing their interrelationships. We analyze histology-to-pathway and pathway-to-histology attention to link histological prototypes with relevant biological pathways. Additionally, we model text-to-pathway and text-to-histology interactions to understand how pathology reports emphasize biological pathways and align with morphological features in WSIs.

To analyze pathology reports, we compute the standard deviation of cross-attention scores across all text segments within a single report to identify key pathways and clusters (Figures.\ref{Figure:Attention}.h,e). Instead of focusing on individual text segments, we consider the entire report to capture the overall diagnostic context. Standard deviation is used instead of averaging attention scores, as it better highlights pathways that receive selective but strong attention from certain segments while being ignored by others, preventing dilution of meaningful signals.

For Prototype 15 ($C=15$), which represents tumor regions and is the most dominant prototype in the WSI, we identify MYC Targets V1, TNFA Signaling via NF-$\kappa$B, and Inflammatory Response as key pathways consistently emphasized by both histology-based and pathology report-based attention (Figures \ref{Figure:Attention}.g,h). These pathways have been shown to be important for prognosis \cite{martin2021pivotal,liang2021identification,lee2015c}. Among these, we visualize the highly attended histological prototypes corresponding to MYC targets V1 and the pathology report, noting that C15 emerges as a highly attended prototype in both (Figures\ref{Figure:Attention}.e,f) This finding underscores strong bidirectional cross attention between the three modalities.

\begin{figure*}[ht]
\includegraphics[width=\linewidth]{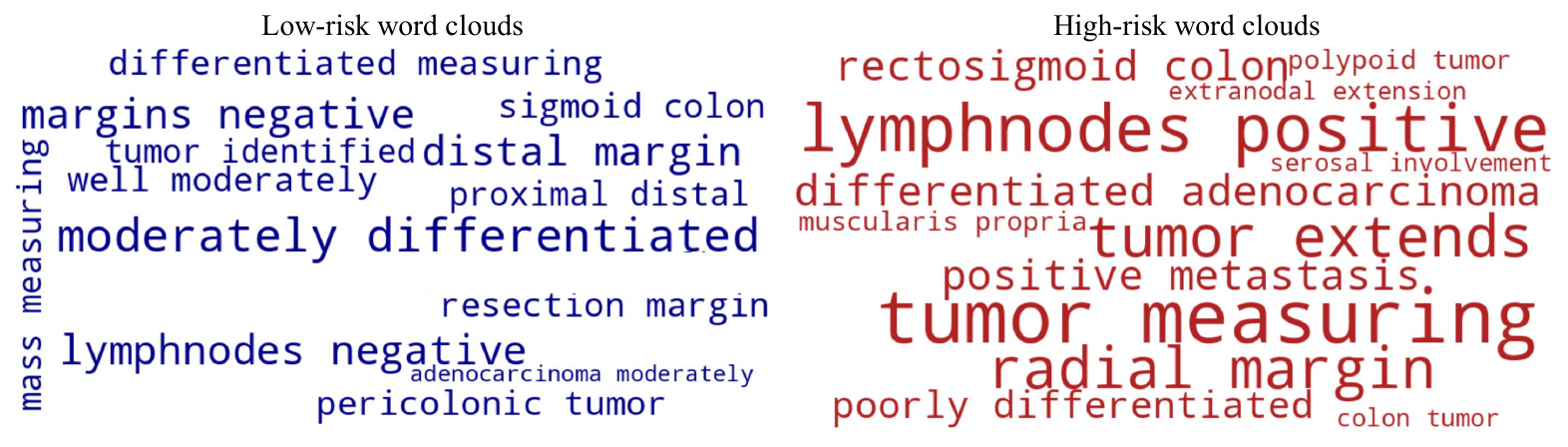}
\caption{Two-phrase wordclouds for high-risk group (red) and low-risk (blue) group for TCGA-CRC depicting words from top text segments based on histology prototypes.}
\label{Figure:wordcloud}
\end{figure*}

\subsection{Word Clouds}
We stratified patients for TCGA-CRC into low- and high-risk groups based on the median cutoff of their predicted risk scores. Using cross-attention mechanisms, we identified the most highly attended text segment within each pathology report, determined by the average attention from all histological prototypes. To explore risk-associated textual patterns, we use the top-ranked text segment for each patient and generated two word clouds—one representing the high-risk group and another for the low-risk group as shown in Fig.\ref{Figure:wordcloud}. The provided word clouds categorize two-word phrases instead of single words. The low-risk word cloud (blue) includes terms like ``margins negative" and ``lymph nodes negative," which indicate that cancer has not spread and are associated with a better prognosis \cite{ogino2010negative}. Additionally, phrases such as ``moderately differentiated" and ``well differentiated" align well with low-risk pathology, as tumors with these characteristics tend to be less aggressive compared to poorly differentiated ones. Conversely, the high-risk word cloud (red) contains terms that indicate advanced disease and poor prognosis, such as ``lymph nodes positive," ``poorly differentiated," ``serosal involvement," and ``radial margin" \cite{shivji2020poorly}. These terms reflect features linked to higher recurrence risk, deeper tissue invasion, and metastatic potential, making them indicators of more aggressive colorectal cancer.

\section{Kaplan-Meier Analysis}
\begin{figure*}[ht]
\centering
\includegraphics[width=0.8\linewidth]{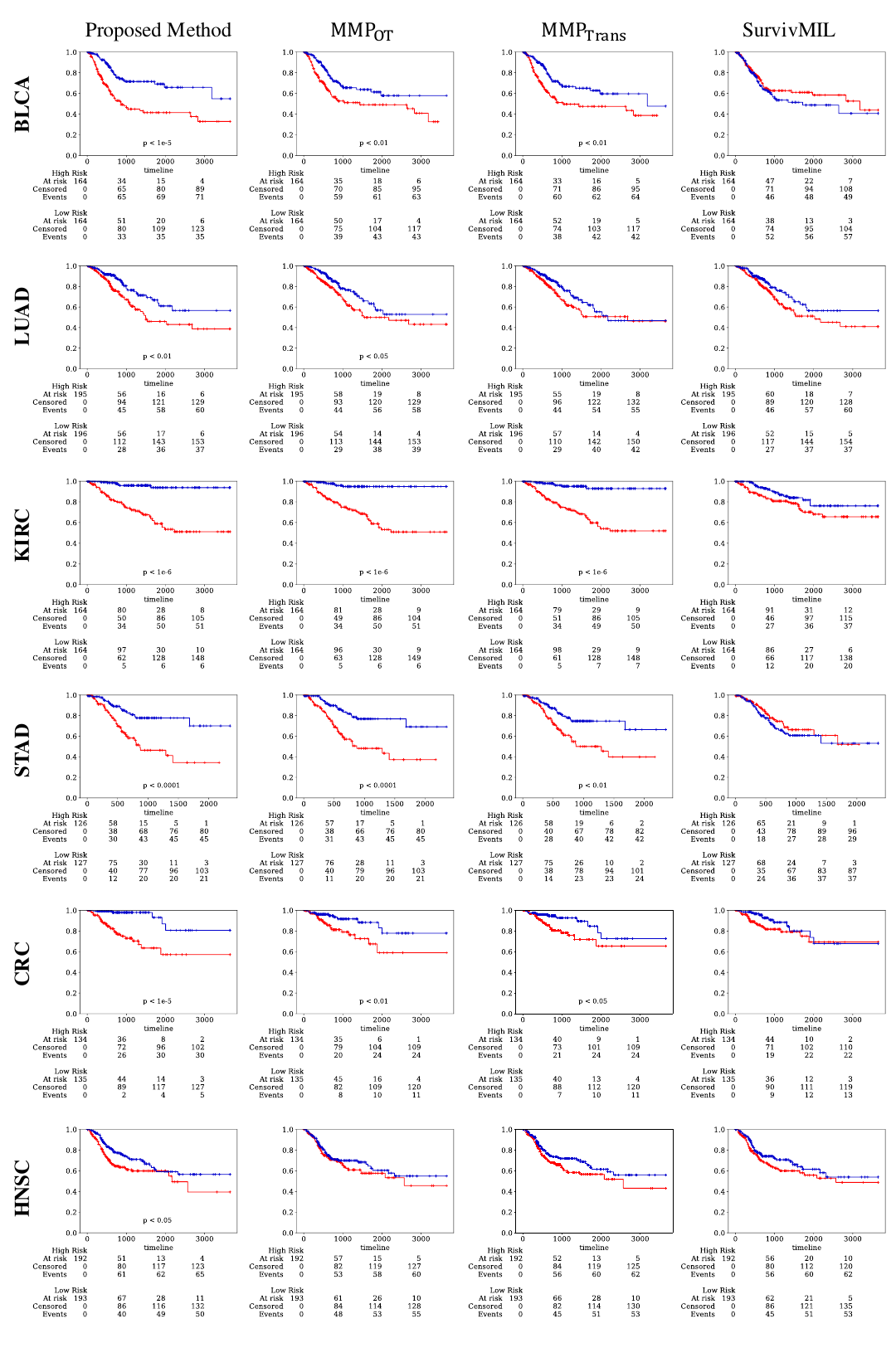}
\caption{Kaplan-Meier curves comparing the proposed method with multimodal baselines. High-risk (red) and low-risk (blue) groups were stratified using the median predicted risk. Statistical significance was assessed using the log-rank test.}
\label{Figure:KM_Curves}
\end{figure*}

Figure.\ref{Figure:KM_Curves} presents Kaplan-Meier survival curves for the predicted high-risk and low-risk groups. Patients with risk scores above the cohort median are classified as high-risk (red), while those below the median are considered low-risk (blue). We compare our proposed model against key baselines, including the best overall multimodal model ($\text{MMP}_{\text{OT}}$), the top transformer-based multimodal baseline ($\text{MMP}_{\text{Trans}}$), and the sole histology-text baseline (SurvivMIL). We use the log-rank test \cite{bland2004logrank} to assess whether the difference between high- and low-risk groups is statistically significant, considering a \textit{p}-value threshold of 0.05.

\end{document}